\documentclass{article}
\PassOptionsToPackage{sort&compress}{natbib}
\usepackage[preprint]{corl_2025}
\AtBeginDocument{\hypersetup{pdftitle={LogLook: Execution-Grounded Memory for Frontier-Model Robot Planners},pdfauthor={Tianyi Jiang, Tianyao Tong, Zhenshan Bing},pdfsubject={},pdfkeywords={Robot memory, foundation-model planners, long-horizon manipulation}}}
\usepackage[T1]{fontenc}
\usepackage{graphicx,booktabs,multirow,xcolor,colortbl,microtype,enumitem,amsmath}
\usepackage[font=small,labelfont=bf]{caption}
\usepackage[capitalise,noabbrev]{cleveref}
\usepackage{newunicodechar}
\newunicodechar{±}{\ensuremath{\pm}}

\definecolor{ours}{HTML}{C0504D}
\definecolor{oursrow}{HTML}{FBECEB}
\newcommand{\astra}{GPT-6 Astra}
\newcommand{\pio}{$\pi_{0.5}$}
\newcommand{\ours}{LogLook}
\setlist{itemsep=1pt,topsep=3pt}

\title{LogLook: Execution-Grounded Memory for Frontier-Model Robot Planners}
\author{
  Tianyi Jiang$^{1}$\quad Tianyao Tong$^{2}$\quad Zhenshan Bing$^{1,*}$\\
  {\small $^{1}$State Key Laboratory for Novel Software Technology and School of Intelligence Science and Technology,}\\
  {\small Nanjing University (Suzhou Campus), Suzhou, China}\\
  {\small $^{2}$School of Computation, Information and Technology, Technical University of Munich, Garching, Germany}\\
  {\small $^{*}$Corresponding author: \texttt{bing@nju.edu.cn}}
}

\begin{document}
\maketitle
\vspace{-1.2cm}
\begin{center}
  \includegraphics[width=\linewidth]{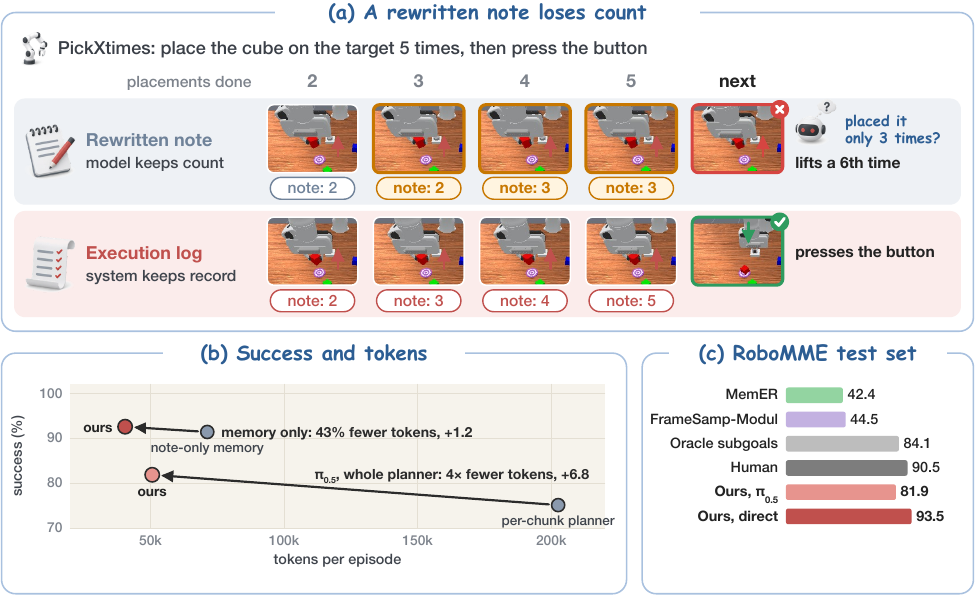}
  \captionof{figure}{\textbf{(a)} The same PickXtimes episode planned by \astra{} with two memories, aligned by the number of placements actually completed. When the model keeps the count in a note that it rewrites at every step, it loses count after the third placement and lifts the cube a sixth time; with the execution log kept by the system, it counts every placement and presses the button. \textbf{(b)} Success against tokens per episode for two paired comparisons on the same episodes: swapping only the memory with the motion planner (256 episodes), and our whole planner against a planner that re-plans after every \pio{} chunk (800 episodes), which also differs in call frequency and actions. \textbf{(c)} Average success on the RoboMME test set; baselines as reported by RoboMME~\citep{dai2026robomme}.}
  \label{fig:teaser}
\end{center}
\vspace{-0.25cm}

\begin{abstract}
Many manipulation tasks require a robot to remember what it has done and what it has seen.
Frontier vision-language models can plan such tasks without task-specific training, but how such a planner should remember an episode is an open question.
We argue that its memory must keep track of what the planner has done and learned, and must let the planner choose what to look at, since every frame costs tokens; a note rewritten at every step meets both poorly.
We propose \ours{}, whose \emph{execution-grounded memory} lets the system keep the record, an execution log and the model's earlier notes, while the model decides where to look: frames are returned only where it expects a change, and demonstrations are skimmed before it inspects the parts it chooses.
Measured with the benchmark's motion planner, \ours{} matches the success of a note-only memory with 43\% fewer tokens and less than half the images, and fails far fewer episodes through what it remembered or saw.
With direct control, \ours{} solves 93.5\% of the RoboMME test set.
\end{abstract}


\section{Introduction}
Many everyday manipulation tasks cannot be solved from the current camera image alone.
A robot asked to place a cube on a target five times must know how many placements it has already made.
A robot asked to lift the container that hides the green cube must have followed the containers while they were shuffled, and one asked to move a cube ``in the same manner as before'' must have noticed whether the demonstrator pushed or carried it.
RoboMME~\citep{dai2026robomme} collects sixteen such tasks, and its evaluation shows that memory is the bottleneck for today's robot policies: the best learned memory reaches about half the success of a \pio{}~\citep{pi2025pi05} executor driven by oracle subgoals.

Frontier vision-language models now perceive and reason well enough to act as robot planners without task-specific training~\citep{gemini2025robotics,zhang2026unexpected,galbot2026astra,liu2026roboicl}, and on RoboMME such a planner already succeeds more often than learned memory policies (\cref{tab:pi05}).
How the planner should remember an episode, however, is still open.
The usual choices trade reliability against cost: a note that the model rewrites at every step, the scratchpad memory of language agents~\citep{yao2022react}, is cheap but has to carry everything the planner will need later, while showing the model every frame keeps all the evidence and makes each decision more expensive as the episode goes on.
The question we ask is what a frontier planner needs to remember, and who should keep it.

We argue that this memory has to meet two requirements.
First, it must keep track of what the planner has done and learned, so that a placement that has just been completed is counted and a colour seen before the containers covered the cubes is still known later; a note that is rewritten at every step can lose either in a single rewrite (\cref{fig:teaser}a), and continuously rewritten memories are known to degrade~\citep{zhang2026faulty}.
Second, it must let the planner choose what to look at: frames recorded during an action rarely change but cost tokens, while some facts, such as a container swap or whether a demonstrator pushed or carried a cube, are visible only for a moment.

\begin{figure}[t]
\centering
\includegraphics[width=\linewidth]{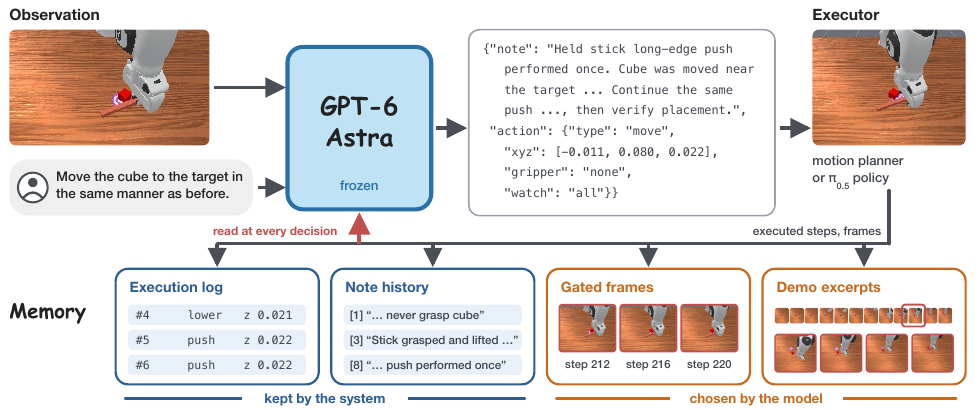}
\caption{\textbf{Overview of execution-grounded memory.} At every decision, \astra{} receives the observation, the instruction and the memory, and returns a note and an action, optionally with image points to watch; the motion planner or \pio{} executes the action. The memory has two parts. The system keeps the record: one log line per executed action with the measured gripper state, and every note the model has written. The model chooses what to see: after an action only the recorded frames that changed near the watched points are returned, and a demonstration is skimmed before the model inspects the ranges it asks for. All snippets come from one MoveCube test episode: the observation, a verbatim (shortened) decision, the resulting log lines and notes, the three frames returned during the watched push, and the demonstration range the model inspected.}
\label{fig:method}
\vspace{-0.3cm}
\end{figure}

Both requirements suggest a division of labour in which the system keeps the record and the model decides where to look, and our execution-grounded memory (\cref{sec:method}, \cref{fig:method}) is built this way.
The system writes a log of every executed action with the gripper state measured right after it, and keeps every note the model has written.
The model, in turn, names the image regions where it expects something to change during an action, and only frames that change there are returned; it skims each demonstration and inspects the parts it chooses.
The design is the same for every task and involves no training.

Measuring what the memory contributes turned out to need an executor whose own failures are rare and can be told apart from the planner's.
With the learned \pio{} executor, swapping only the memory leaves the success rate unchanged, because empty grasps, stalls and late button presses decide almost all episodes in which the two memories disagree.
With the benchmark's motion planner, whose execution fails far less often, our memory reaches the success of a note-only memory while sending the model 43\% fewer tokens and less than half as many images. Separating the failures shows why the success rates end up close: the note-only memory fails three times as many episodes for reasons other than motion planning (12 against 4), and at least 11 of its 12, against one of our 4, come from what the planner remembered or saw; motion-planning failures, which the benchmark counts as failed episodes, strike both memories at random and decide most of the remaining episodes.
With direct control our planner solves 93.5\% of the RoboMME test set, and over the learned executor it comes close to oracle subgoals while using a quarter of the tokens of a planner that re-plans after every action chunk.
In summary, our contributions are as follows:
\begin{itemize}
  \item We identify two requirements for the memory of a frontier-model robot planner, keeping track of what was done and learned and choosing what to look at, and show in a per-episode analysis that a rewritten note fails on both.
  \item We propose an execution-grounded memory in which the system keeps a record of what was executed and measured while the model decides which frames and which parts of a demonstration to look at.
  \item We show that a learned executor hides differences between memories, and that with the benchmark's motion planner our memory matches the success of a note-only memory at well under half the image cost, with far fewer failures caused by memory and observation.
\end{itemize}

\section{Related Work}
\paragraph{Memory in robot manipulation policies.}
Most robot policies, from vision-language-action models such as OpenVLA~\citep{kim2024openvla}, $\pi_0$~\citep{black2024pi0} and \pio{}~\citep{pi2025pi05} to their predecessors, act on the current observation or a short window.
Memory has been added as a bank of visual features (SAM2Act~\citep{fang2025sam2act}), a perceptual-cognitive memory (MemoryVLA~\citep{shi2025memoryvla}), state-change tokens (EvoMem-VLA~\citep{na2026evomem}), or keyframes selected by a high-level VLM that issues subtasks to \pio{} (MemER~\citep{sridhar2025memer}).
RoboMME~\citep{dai2026robomme} compares these designs and several of its own (frame sampling, token dropping, recurrent and test-time-trained memory) under one protocol, and RMBench~\citep{chen2026rmbench} studies memory-dependent manipulation on a second suite; both report large gaps to oracle or human performance.
All of these methods learn the memory jointly with the policy. We study the complementary setting in which a frozen frontier model plans and the question is how its memory should be organised.

\paragraph{Foundation models as robot planners.}
Language and vision-language models have planned robot behaviour by scoring skills~\citep{ahn2022saycan}, with closed-loop textual feedback~\citep{huang2022innermonologue}, by writing code~\citep{liang2022cap}, and by composing spatial constraints or value maps for a motion planner~\citep{huang2023voxposer,huang2024rekep,liu2024moka}.
Hierarchical systems pair a VLM that issues language subtasks with a VLA that executes them~\citep{shi2025hirobot,pi2025pi05,sridhar2025memer}, and frontier models are now evaluated directly as zero-shot policies~\citep{gemini2025robotics,zhang2026unexpected,galbot2026astra,jia2026agentpolicy}.
Closest to us, RoboICL~\citep{liu2026roboicl} gives \astra{} demonstration context and an interaction memory of observation-action-receipt records, and Harness VLA~\citep{zhang2026harnessvla} steers a frozen VLA with a memory-guided agent.
These works use memory as one part of a larger system; we isolate it, on a benchmark built to require it, and show which records a frontier planner needs kept for it.

\paragraph{Frontier models on RoboMME.}
Two recent systems apply foundation-model planners to RoboMME.
Chen et~al.~\citep{chen2026astrarobomme} call \astra{} as a subtask planner for a RoboMME-trained \pio{} only when a learned completion monitor reports that a subtask has finished, using task-specific prompts.
Ledger~\citep{dieudonne2026ledger} gives a Claude planner an external object record built by a visual tracker and pairs it with a re-trained \pio{}.
They decide when the planner is called or what is tracked for it; we study how the planner's own memory should be organised.

\paragraph{Memory and active perception in language agents.}
Language agents keep memory as interleaved reasoning traces~\citep{yao2022react}, reflections~\citep{shinn2023reflexion}, retrieved records of experience~\citep{park2023generative,packer2023memgpt}, or skill libraries~\citep{wang2023voyager}; continuously rewriting such a memory with an LLM degrades it~\citep{zhang2026faulty}.
Agents for long videos and interactive worlds select which frames to look at instead of processing all of them~\citep{wang2024videoagent,han2026vista}.
Unlike these agents, a robot planner changes the world itself, so the facts it most needs are its own actions and their physical outcome, and it receives tens of frames per second, so it has to decide what to look at before or while something happens. Our memory is designed around these two differences.

\section{Method}\label{sec:method}
We first state the problem and the two requirements on the memory (\cref{sec:formulation,sec:problems}); the remaining subsections describe our memory.

\subsection{Problem formulation}\label{sec:formulation}
We consider a frozen vision-language model that plans an episode through a sequence of decisions.
At each decision it receives the current observation (front and wrist images, end-effector pose, gripper opening and environment step), the task instruction, a task note and a memory, and returns a free-text note and an action, which an executor carries out.
Everything the planner knows about earlier parts of the episode, including the demonstration, reaches it only through the memory.
The question we study is what this memory should contain and who should write it, so that the planner succeeds while being shown as few tokens and images as possible.

\subsection{Two requirements on the memory}\label{sec:problems}
The most common memory for a language-model agent is a note that the model rewrites at every step~\citep{yao2022react}.
In a robot episode this note has to do two jobs.

\paragraph{Keeping track of what was done.}
If the note is the only record of the episode, every rewrite can change it: a repetition that has just been completed may not be counted, and a fact recorded earlier, such as which container covers which cube, may disappear a few rewrites later.
Continuously rewritten memories are known to degrade in this way~\citep{zhang2026faulty}.
The record of what was executed and what was learned should therefore not depend on every rewrite getting it right.

\paragraph{Choosing what to look at.}
Some facts are visible only for a moment, during an action or in a demonstration, yet every image the planner is shown costs tokens.
Showing every frame recorded during an action, or a whole demonstration at a fixed stride, is expensive although in most tasks nothing moves unless the robot moves it, and a fixed stride can still miss the moment that matters, such as whether a demonstrator pushed or carried a cube.
Showing none would hide container swaps and moving cubes.
The planner can predict where and when something is likely to change, so we let it decide what it is shown.

\subsection{Execution-grounded memory}\label{sec:egm}
The memory consists of an execution log and the history of earlier notes, both kept by the system, and of frames and demonstration excerpts that the model asks to see.

The planner is a loop of model calls (\cref{fig:method}); selected frames and notes of one episode are shown in \cref{fig:timeline}.
At each decision \astra{} receives the front and wrist images, the end-effector pose and gripper opening, the current environment step, the task instruction and task note, the result of its previous action and the memory described below, and returns a JSON object with a free-text note and one action.
Actions are \texttt{locate} (pixels to 3-D points; no motion), \texttt{move} (a waypoint with gripper command and optional start or arrival time), \texttt{wait}, and \texttt{sequence}, a chain of up to six moves or waits that runs without further model calls.
Each step of a sequence may carry a check on the gripper opening or end-effector height (for example, stop if the grasp closed on nothing); the sequence stops at the first failed check, and the model is asked again.
With the learned executor, \texttt{move} is replaced by \texttt{delegate}: a subgoal phrase, a target pixel and a number of 16-step chunks for \pio{} to execute, followed by a per-chunk trace of the lowest end-effector height and the finger opening.

\subsection{Keeping track of what was done}
\paragraph{Execution log.}
So that completed actions are not lost in a rewrite, the system appends one line for every action that is executed: its index, the environment step, the command, and the finger opening and end-effector height measured right after it, or \texttt{NOT EXECUTED} if the motion planner rejected it.
The full log is shown at every decision.
It records which commands were carried out and what the gripper measured afterwards, without depending on the model having written it down; whether a command achieved its purpose, for example whether the intended cube was grasped, the model still judges from this record and the images.
The note-only memory shows only the last eight actions.

\paragraph{Note history.}
The model still writes a note at each decision, and still decides what goes in it: the identity of the target, the reading of the demonstration, its plan.
In addition to the latest note, it now sees all of its earlier notes verbatim, so a fact that a rewrite dropped can be recovered.

\subsection{Choosing what to look at}
\paragraph{Expectation-gated frames.}
While an action runs, the front camera is recorded about every four environment steps; for long actions the stride grows so that at most 16 evenly spaced frames, always including the last, are kept.
To avoid sending all of them, we let the model predict where something may change: an action may carry \texttt{watch} with a list of front-image points, or \texttt{"all"}.
After the action, the system returns only the recorded frames in which the image changed within a small circle (28 px at 512 px resolution) around a watched point: a frame and its predecessor are returned when more than 5\% of the pixels inside the circles differ between them by more than 30 intensity levels. The system also says how many frames it withheld.
Without \texttt{watch}, no intermediate frames are returned.
The note-only memory returns every recorded frame.

\paragraph{Skim, then inspect.}
Demonstrations are shown before execution in segments.
The model first sees every 8th frame of a segment and writes a running summary of the demonstration; it may then ask for closer looks at frame ranges of its choice, which are shown at every 2nd frame (at most 24 frames) before the next segment.
The model chooses the ranges; the system applies the fixed strides and caps above. The running summary is itself a text state that the model rewrites, but it is written while the evidence is in view, and frames can be inspected again before it is final. The final summary is given to the planner as its initial note when execution starts.
The note-only memory shows every 2nd frame (every 4th for long videos) of the whole demonstration.

\section{Experiments}\label{sec:exp}
Our experiments ask three questions.
How does our memory change what the planner gets wrong and what it costs, once the executor's own failures are taken out of the comparison (\cref{sec:measure})?
What does each component contribute (\cref{sec:components})?
And how does the resulting planner perform on the full benchmark (\cref{sec:full})?

\subsection{Experimental setup}\label{sec:setup}
\paragraph{Benchmark.}
RoboMME~\citep{dai2026robomme} contains 16 tabletop tasks for a Franka Panda in simulation, grouped into four suites of four tasks: \emph{Counting} (BinFill, PickXtimes, SwingXtimes, StopCube), \emph{Permanence} (VideoUnmask, ButtonUnmask, VideoUnmaskSwap, ButtonUnmaskSwap), \emph{Reference} (PickHighlight, VideoRepick, VideoPlaceButton, VideoPlaceOrder) and \emph{Imitation} (MoveCube, InsertPeg, PatternLock, RouteStick).
Nine tasks (the five Video tasks and the four Imitation tasks) begin with a demonstration video, some of more than a thousand frames.
The test split has 50 episodes per task (800 in total) at three difficulty levels; \cref{fig:gallery} shows one episode of each task.
Episodes end on success, on failure as judged by the benchmark, or after 1300 environment steps.

\paragraph{Executors.}\label{sec:executors}
We use the planner with two executors.
\emph{Learned executor}: the RoboMME \pio{} checkpoint trained on grounded subgoals (``pick up the green cube at $\langle x,y\rangle$''), the same executor that RoboMME evaluates with oracle subgoals (84.1\%), which we refer to as the oracle-subgoal result.
The planner issues a subgoal phrase from the executor's training vocabulary together with a target pixel and the number $k\in\{1,\dots,8\}$ of 16-step action chunks to run; it may also wait or query 3-D points.
\emph{Direct control}: the planner issues end-effector waypoints that the benchmark's motion planner executes.
It may query the 3-D world point under any pixel of the front or wrist image, computed from the depth image and camera calibration; it never sees simulator object state.
As in RoboMME's evaluation of learned policies, the simulation waits while the planner is queried. We keep RoboMME's rule that an episode fails when executing an action raises an error, which with direct control happens when the motion planner finds no path to the commanded pose; our planner has no ``done'' action, and success is decided only by the environment. Direct control removes executor failures that the planner cannot influence (\cref{sec:measure}) and is our main setting for measuring memory; the learned executor is the setting comparable to prior learned policies.

\paragraph{Baselines.}
We compare against (1) the \emph{note-only memory}, which uses the same code, prompts, executor, task notes and budgets as ours and replaces the four components by their simplest form, namely the model's latest note, the last eight actions, every frame recorded during an action, and the whole demonstration at a fixed stride;
(2) the \emph{per-chunk planner}, which calls the model after every 16-step chunk of \pio{}, asks it to rewrite a structured state description and otherwise uses the note-only memory, so that it also differs from ours in how often the model is called and in the actions available to it;
(3) the learned memory policies of RoboMME~\citep{dai2026robomme}, all built on \pio{}, of which FrameSamp-Modul, which feeds \pio{} sampled past frames through a modulation module, and MemER~\citep{sridhar2025memer}, which selects keyframes with a fine-tuned VLM that issues subgoals to \pio{}, are the strongest;
and (4) GroundSG, \pio{} trained to execute grounded subgoals, driven by subgoals from simulator ground truth (``oracle subgoals'') or predicted by a fine-tuned Qwen3-VL.
\Cref{tab:interface} summarises what each system observes and controls.

\begin{table}[h]
\centering\footnotesize
\caption{What each compared system decides with, how its actions are executed, which inputs it receives beyond the RGB images and proprioception, and which parts are trained on RoboMME.}
\label{tab:interface}
\resizebox{\linewidth}{!}{\begin{tabular}{@{}l l l l l@{}}
\toprule
System & Decision maker & Executor (action) & Beyond RGB + proprio. & Trained on RoboMME \\
\midrule
MemER, FrameSamp-Modul & learned memory policy & \pio{} (joint actions) & -- & memory and executor \\
GroundSG, oracle subgoals & simulator ground truth & \pio{} (joint actions) & privileged subgoals & executor \\
Human~\citep{dai2026robomme} & human player & oracle skill interface & privileged skills & -- \\
\midrule
\rowcolor{oursrow}Ours, \pio{} executor & \astra{} & \pio{} (subgoal phrases) & depth, calibration, task notes & executor only (same checkpoint) \\
\rowcolor{oursrow}Ours, direct control & \astra{} & motion planner (waypoints) & depth, calibration, task notes & nothing \\
\bottomrule
\end{tabular}
}
\end{table}

\paragraph{Task notes.}
With direct control, each task family has a short note stating what the instruction means and when the benchmark counts each step as done. These conditions are taken from the benchmark's success-checking code, for example that a press counts only once the button is pushed down by at least 5\,mm. Because the planner cannot observe such quantities directly, the note also states what they correspond to in its own observations, such as the fingertip heights that real presses reached, measured on training episodes. Without this information the planner has to guess how far a step must be carried out. The notes contain no episode-specific information such as object positions or repetition counts. With the learned executor we use notes written for the \pio{} subgoal interface, which also describe how the executor's phrases are used and give some procedural hints, for example to hover above the button before pressing it; the per-chunk planner uses the same notes. The observation values in the notes were measured offline on training-split episodes driven by oracle subgoals; no test episode and no simulator state is used at test time. In every comparison of two memories the notes are identical, and \cref{sec:full} measures what the notes contribute.

\paragraph{Metrics and protocol.}
We report the success rate, the tokens per episode, the number of images sent to the model per episode and the number of model calls.
Each configuration is run once per episode with \astra{}, with a budget of 1300 environment steps and 1.5\,M tokens per episode. Further settings are listed in \cref{app:impl}.
Cost is measured in tokens, counted as the input and output tokens of the planner's own requests, and in the number of images sent to the model, both including demonstration viewing.
For the RoboMME methods we quote the published numbers~\citep{dai2026robomme}.
Whenever we compare two planners on the same episodes we use a two-sided sign test over the episodes in which they disagree, and we call a difference significant at the 5\% level; per-comparison counts are given in the tables.
With direct control we also split the failed episodes into motion-planning failures and all other failures, which come from what the planner decided, remembered or saw. The motion planner of RoboMME occasionally finds no path even for ordinary moves, such as a descent onto a cube from a few centimetres above it; this happens in about 2--12\% of the episodes of every configuration we ran, with no significant dependence on the memory.

\begin{figure}[t]
  \centering
  \includegraphics[width=\linewidth]{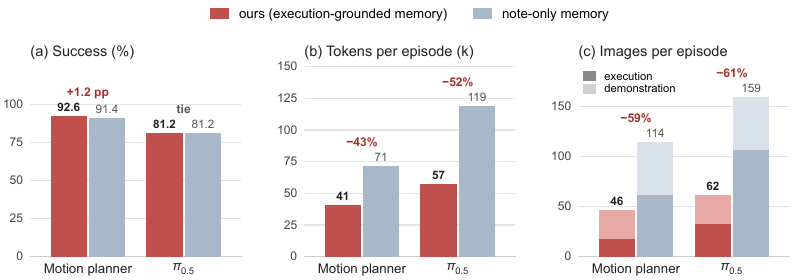}
  \caption{Execution-grounded memory against note-only memory, with everything else fixed. With either executor the two memories reach similar success: with the learned \pio{} executor, executor failures decide most episodes in which they disagree, and with the motion planner, motion-planning failures do. In both settings our memory sends the model far fewer tokens and images; in (c), the lighter part of each bar counts demonstration frames and the darker part frames from execution.}
  \label{fig:ablation}
  \vspace{-0.3cm}
\end{figure}

\begin{table}[t]
\centering\small
\caption{Execution-grounded vs.\ note-only memory (same code, prompts, executor, task notes and budgets). Motion planner: test episodes 4--19 of every task; \pio{}: episodes 4--11. Failures are split into motion-planning failures and all other failures.}
\label{tab:ablation}
\resizebox{\linewidth}{!}{\begin{tabular}{l l r r r r r}
\toprule
Executor (episodes) & Memory & Success (\%) & Failures: planning / other & Ours wins / losses & Tokens / ep. & Calls / ep. \\
\midrule
\multirow{2}{*}{Motion planner (256)} & \textbf{execution-grounded} & \textbf{237/256 (92.6)} & 15 / 4 & \multirow{2}{*}{19 / 16} & \textbf{40.5k} & \textbf{7.3} \\
 & note-only & 234/256 (91.4) & 10 / 12 & & 71.3k & 9.2 \\
\midrule
\multirow{2}{*}{$\pi_{0.5}$ (128)} & \textbf{execution-grounded} & \textbf{104/128 (81.2)} & -- & \multirow{2}{*}{7 / 7} & \textbf{57.4k} & \textbf{11.5} \\
 & note-only & 104/128 (81.2) & -- & & 118.6k & 12.0 \\
\bottomrule
\end{tabular}
}
\vspace{-0.3cm}
\end{table}

\subsection{Controlled memory comparison}\label{sec:measure}
\paragraph{With the learned executor.}
We first swap the memory with the learned \pio{} executor and change nothing else.
Both memories solve the same number of episodes (\cref{fig:ablation}a).
Of the fourteen episodes in which they disagree, twelve are decided by the executor, through empty grasps, stalls that run out the step budget and button presses that come too late or do not register; in two RouteStick episodes, replaying the identical command sequence even succeeds. \Cref{tab:attrib} lists every discordant episode with its cause, and \cref{fig:failures} shows examples.
With an executor that fails this often, the comparison cannot show what the memory contributes, so we measure it with the benchmark's motion planner, whose own failures are much rarer.

\begin{figure}[t]
  \centering
  \includegraphics[width=\linewidth]{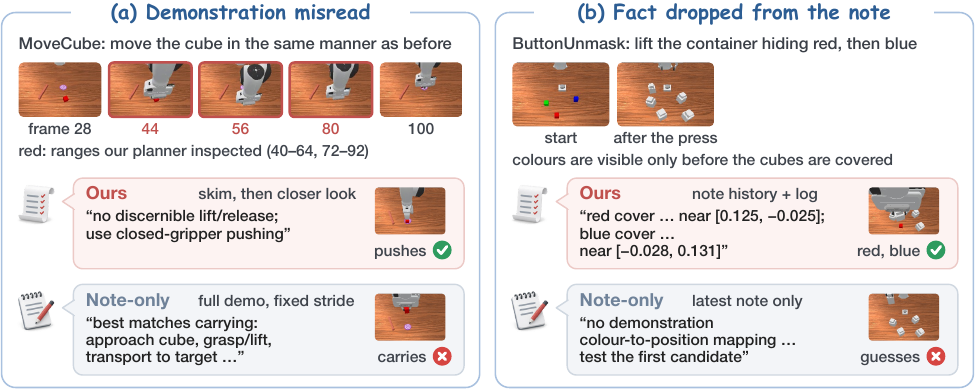}
  \caption{Two failures of the note-only memory, same episode under both memories (direct control). Quotes are verbatim excerpts of the planner's notes. \textbf{(a)} MoveCube: in the demonstration the cube slides under the closed gripper. Our planner skims the demonstration, inspects frames 40--64 and 72--92 closely and reads a push; with the full demonstration at a fixed stride, the note-only memory reads a carry. \textbf{(b)} ButtonUnmask: the colours are visible only before the containers cover the cubes, and the later notes of the note-only memory no longer contain them.}
  \label{fig:qual}
  \vspace{-0.3cm}
\end{figure}

\paragraph{With the motion planner.}\label{sec:swap}
On twice as many episodes, the two memories again reach similar success, 92.6\% against 91.4\% of the 256 episodes (\cref{tab:ablation}), but the execution-grounded memory is much cheaper, sending the model 41k instead of 71k tokens and 46 instead of 114 images per episode (\cref{fig:ablation}b,c).
The failures, however, are of a different kind. The note-only memory fails 12 episodes for reasons other than motion planning, the execution-grounded memory 4; motion-planning failures end 10 and 15 episodes, respectively, and they occur at the same low rate in every configuration we ran (\cref{sec:setup}). They therefore decide most episodes in which the two memories disagree and mask the difference in what the planner remembers.
\paragraph{Why the note-only memory fails.}
We read every episode that only one of the two memories solves, decision by decision, and checked it against the executed-step feedback (\cref{tab:attrib}).
Of the 19 episodes that only our memory solves, the note-only planner loses five to motion-planning failures and all but one of the other fourteen to what it remembered or saw.
In five, a repetition or a deposit is executed completely but the next note does not count it, and the planner performs one too many (\cref{fig:teaser}a).
In three, a fact recorded earlier, such as which container covers which cube or where a swapped container went, disappears a few rewrites later and the planner has to guess (\cref{fig:qual}b).
In five, the demonstration is misread: a push is read as a carry, the wrong peg or the wrong end of a peg is taken, or a route is reversed (\cref{fig:qual}a).
Of the 16 episodes that only the note-only memory solves, our planner loses 13 to motion-planning failures and three to its own errors: a demonstration it read as a carry despite a closer look, a StopCube press that came too late and a slipped grasp.

\begin{table}[t]
\centering\small
\caption{Switching off one memory component at a time, motion planner, hard test episodes (7, 11, 15 and 19 of every task). Failures are split into motion-planning failures and all other failures. The skim-then-inspect demonstration viewer only matters for tasks with a demonstration and is evaluated on those. Tokens, images and calls are per episode; images are split into frames seen during execution (including the current front and wrist images) and demonstration frames. ``Gated frames'': expectation-gated frames.}
\label{tab:components}
\resizebox{\linewidth}{!}{\setlength{\tabcolsep}{4pt}\begin{tabular}{l r r r r r r r}
\toprule
Memory & Success & Failures: planning / other & Full wins / losses & Tokens & \multicolumn{2}{c}{Images (exec.\ / demo)} & Calls \\
\midrule
\multicolumn{8}{@{}l}{\textit{All 16 tasks, 64 episodes}} \\
\rowcolor{oursrow}Ours (full) & \textbf{56/64} & 8 / 0 & -- & 50.3k & 19 & 30 & 6.7 \\
\quad without execution log & 58/64 & 3 / 3 & 6 / 8 & 40.0k & 21 & 30 & 7.4 \\
\quad without note history & 61/64 & 1 / 2 & 2 / 7 & 45.8k & 23 & 29 & 8.0 \\
\quad without gated frames & 58/64 & 4 / 2 & 5 / 7 & 61.1k & 83 & 28 & 7.6 \\
Note-only memory & 53/64 & 4 / 7 & 10 / 7 & 92.6k & 90 & 52 & 8.7 \\
\midrule
\multicolumn{8}{@{}l}{\textit{The 9 tasks with a demonstration, 36 episodes}} \\
\rowcolor{oursrow}Ours (full) & \textbf{31/36} & 5 / 0 & -- & 50.7k & 12 & 53 & 5.9 \\
\quad without skim-then-inspect & 32/36 & 0 / 4 & 2 / 3 & 46.3k & 10 & 93 & 5.1 \\
Note-only memory & 32/36 & 1 / 3 & 4 / 5 & 97.1k & 63 & 93 & 7.9 \\
\bottomrule
\end{tabular}
}
\vspace{-0.3cm}
\end{table}

\subsection{What each component contributes}\label{sec:components}
\begin{figure}[t]
  \centering
  \includegraphics[width=\linewidth]{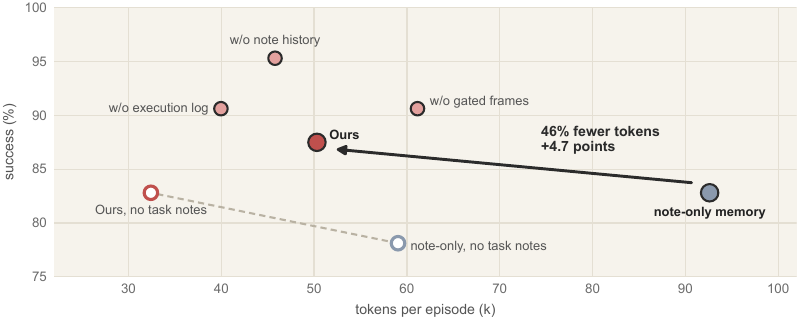}
  \caption{Success against tokens per episode on the hard episodes, for our memory, our memory with one component switched off, and the note-only memory, each with and without task notes. Success counts motion-planning failures as failures, which differ by a few episodes between configurations at random (\cref{tab:components} splits the failures). The arrow marks replacing the note-only memory by ours; the dashed line the same without task notes. Skim-then-inspect only applies to tasks with a demonstration and is shown in \cref{tab:components}.}
  \label{fig:tradeoff}
  \vspace{-0.3cm}
\end{figure}
We switch the components off one at a time on the hard episodes, that is, all episodes among 4--19 that RoboMME labels as hard (7, 11, 15 and 19 of every task) (\cref{tab:components,fig:tradeoff}).
Success alone does not separate the components here: motion-planning failures end one to eight episodes of every configuration, more than any component changes. We therefore look at the other failures. With all four components the planner fails none of the 64 episodes for any reason other than motion planning; without any one component it fails two to four, and with the note-only memory seven. Each single-component difference rests on only two to four episodes, so we read these results as consistent with the failure analysis rather than as separate proof for each component; the clear effect of each component is on cost.

\paragraph{Keeping track of what was done.}
Without the execution log the planner fails two PickXtimes episodes and one SwingXtimes episode, all of them counting tasks.
Without the note history it fails an InsertPeg and a VideoUnmaskSwap episode.
Both components are text and add about 10k and 4k tokens per episode; they cost tokens rather than save them.

\paragraph{Choosing what to look at.}
Expectation-gated frames account for most of the savings: without them the planner sees 83 instead of 19 execution images per episode and uses a fifth more tokens, and it also fails a StopCube and a VideoRepick episode.
The requests follow the dynamics of each task.
In tasks where nothing moves unless the robot moves it, such as BinFill or PickXtimes, the planner almost never asks for intermediate frames; in StopCube and ButtonUnmaskSwap, where it must follow continuous motion, it asks in most decisions and usually wants every frame; in PickHighlight and ButtonUnmask, where changes happen at known places, it mostly names the points to watch.

\begin{figure}[t]
  \centering
  \includegraphics[width=\linewidth]{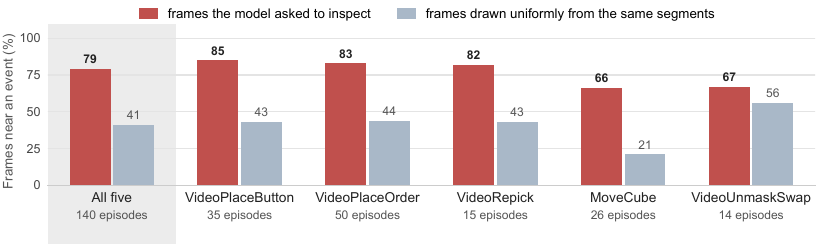}
  \caption{Fraction of frames within eight frames of a ground-truth demonstration event, for the frames the model asked to inspect and for frames drawn uniformly from the same skimmed segments. All direct-control test episodes with at least one closer look, in the five tasks whose demonstrations contain discrete object events.}
  \label{fig:closer}
  \vspace{-0.3cm}
\end{figure}

Without skim-then-inspect the planner fails four of the 36 demonstration episodes for reasons other than motion planning, among them a MoveCube demonstration read as a carry and two RouteStick episodes; with it, none.
Skimming and inspecting halves the demonstration images, though not the tokens, because every segment and closer look is a separate call.
To check whether closer looks are well placed, we replayed the demonstrations in simulation and marked the ground-truth events, such as an object starting or stopping to move or a button being pressed.
The frames the model asks to inspect are about twice as likely to lie near an event as frames drawn uniformly from the same segments (\cref{fig:closer}), and the effect holds for every task except VideoUnmaskSwap, whose demonstration consists mostly of containers in motion.

\subsection{The full benchmark}\label{sec:full}

\begin{figure}[t]
  \centering
  \includegraphics[width=\linewidth]{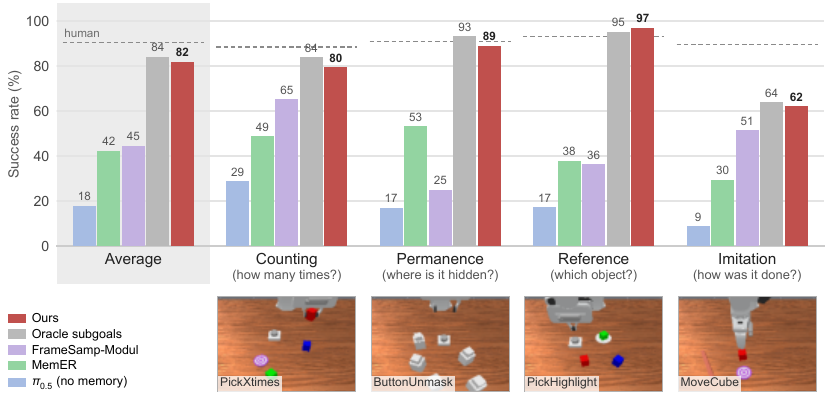}
  \caption{Success on the RoboMME test set, averaged and per suite, for \pio{}-based systems: MemER, oracle subgoals and our planner drive the same RoboMME \pio{} checkpoint, while \pio{} without memory and FrameSamp-Modul are separately trained \pio{} policies. Baselines are the averages published by RoboMME~\citep{dai2026robomme}; the dashed line is human performance, obtained with RoboMME's oracle skill interface. SAM2Act+ and GroundSG with Qwen3-VL subgoals are listed in \cref{tab:main}. Below: frames from our episodes.}
  \label{fig:results}
  \vspace{-0.2cm}
\end{figure}

\begin{table}[t]
\centering\footnotesize
\caption{Per-task success (\%) on the RoboMME test set (50 episodes per task). Baselines as reported by RoboMME~\citep{dai2026robomme}; bold: best per column excluding humans. $^\ddagger$Subgoals from simulator ground truth.}
\label{tab:main}
\resizebox{\linewidth}{!}{\setlength{\tabcolsep}{2.6pt}
\begin{tabular}{lccccccccccccccccc}
\toprule
 & \multicolumn{4}{c}{Counting} & \multicolumn{4}{c}{Permanence} & \multicolumn{4}{c}{Reference} & \multicolumn{4}{c}{Imitation} & \\
\cmidrule(lr){2-5}\cmidrule(lr){6-9}\cmidrule(lr){10-13}\cmidrule(lr){14-17}
Method & \shortstack{Bin\\Fill} & \shortstack{Pick\\Xtimes} & \shortstack{Swing\\Xtimes} & \shortstack{Stop\\Cube} & \shortstack{Video\\Umsk} & \shortstack{Button\\Umsk} & \shortstack{Video\\UmskS} & \shortstack{Button\\UmskS} & \shortstack{Pick\\HighL} & \shortstack{Video\\Repick} & \shortstack{Video\\PlcBtn} & \shortstack{Video\\PlcOrd} & \shortstack{Move\\Cube} & \shortstack{Insert\\Peg} & \shortstack{Pattern\\Lock} & \shortstack{Route\\Stick} & Avg. \\
\midrule
Human & 96.0 & 100.0 & 80.0 & 78.0 & 90.0 & 92.0 & 92.0 & 90.0 & 92.0 & 92.0 & 98.0 & 90.0 & 90.0 & 98.0 & 84.0 & 86.0 & 90.5 \\
$\pi_{0.5}$ (no memory) & 30.0 & 42.9 & 35.6 & 6.7 & 20.4 & 22.2 & 18.7 & 6.7 & 11.3 & 0.4 & 31.1 & 25.8 & 26.0 & 1.6 & 2.9 & 4.7 & 17.9 \\
SAM2Act+ & 40.0 & 76.0 & 25.3 & 0.0 & 27.3 & 32.0 & 18.0 & 26.7 & 17.3 & 5.3 & 24.7 & 20.0 & 29.3 & 0.0 & 0.0 & 0.0 & 21.4 \\
MemER & 56.7 & 79.3 & 59.3 & 0.0 & 81.3 & 72.0 & 38.0 & 21.3 & 70.7 & 25.3 & 30.0 & 26.0 & 82.7 & 6.7 & 16.7 & 12.0 & 42.4 \\
GroundSG, Qwen3-VL sg. & 52.0 & 92.7 & 7.3 & 0.0 & 88.7 & 24.0 & 30.7 & 14.0 & 15.1 & 25.3 & 54.0 & 31.8 & 71.6 & 3.3 & 6.7 & 6.0 & 32.7 \\
FrameSamp-Modul & 39.6 & 87.3 & 92.0 & 42.0 & 32.7 & 25.1 & 24.4 & 18.2 & 22.9 & 30.4 & 60.0 & 32.0 & 77.8 & 7.6 & 53.6 & 66.7 & 44.5 \\
GroundSG, oracle sg.$^\ddagger$ & 85.8 & \textbf{100.0} & \textbf{100.0} & 49.7 & 98.8 & 95.0 & \textbf{99.2} & 80.2 & 83.3 & 97.3 & \textbf{100.0} & \textbf{100.0} & 87.8 & 15.6 & \textbf{97.0} & 55.6 & 84.1 \\
\midrule
\rowcolor{oursrow}\textbf{Ours} ($\pi_{0.5}$ executor) & 88.0 & 98.0 & \textbf{100.0} & 32.0 & \textbf{100.0} & \textbf{96.0} & 94.0 & 66.0 & 90.0 & \textbf{98.0} & \textbf{100.0} & \textbf{100.0} & \textbf{90.0} & 12.0 & 86.0 & 60.0 & 81.9 \\
\rowcolor{oursrow}\textbf{Ours} (direct control) & \textbf{100.0} & 92.0 & 80.0 & \textbf{84.0} & \textbf{100.0} & \textbf{96.0} & 94.0 & \textbf{96.0} & \textbf{94.0} & 96.0 & 98.0 & 98.0 & 86.0 & \textbf{92.0} & 94.0 & \textbf{96.0} & \textbf{93.5} \\
\bottomrule
\end{tabular}
}
\vspace{-0.2cm}
\end{table}

\begin{table}[t]
\centering\small
\caption{\pio{}-based systems on all 800 test episodes. MemER, oracle subgoals, the per-chunk planner and ours drive the same RoboMME \pio{} checkpoint; FrameSamp-Modul is a separately trained \pio{} policy. ``Official run'': per-episode results released with RoboMME, which can differ slightly from the published averages in \cref{tab:main}. ``Oracle subgoals'': subgoals from simulator ground truth, our run of RoboMME's evaluation code (RoboMME reports 84.1\%). Wins and losses count the episodes in which only one of the two planners succeeds.}
\label{tab:pi05}
\resizebox{\linewidth}{!}{\begin{tabular}{l l r r r r}
\toprule
Planner & Memory & Success (\%) & Ours wins / losses & Tokens / ep. & Calls / ep. \\
\midrule
MemER (official run) & keyframes & 341/800 (42.6) & 327 / 13 & -- & -- \\
FrameSamp (official run) & sampled frames & 372/800 (46.5) & 327 / 44 & -- & -- \\
Oracle subgoals & simulator & 658/800 (82.2) & 59 / 62 & -- & -- \\
Per-chunk planner & state note, rewritten & 601/800 (75.1) & 89 / 35 & 203k & 22.8 \\
\midrule
\rowcolor{oursrow}\textbf{Ours} & \textbf{execution-grounded} & \textbf{655/800 (81.9)} & -- & \textbf{51k} & \textbf{10.5} \\
\bottomrule
\end{tabular}
}
\vspace{-0.3cm}
\end{table}

\paragraph{With the learned executor.}
Even without our memory, a frontier planner succeeds more often than the learned memory policies.
On the 800 test episodes (\cref{fig:results,tab:pi05}), the per-chunk planner solves 75.1\%, against 42.6\% for the official MemER run.
Our full planner improves on this significantly, to 81.9\%, with a quarter of the tokens; this gain comes from the whole system, which also differs in how often the model is called and in the actions it can take.
The planner is now close to oracle subgoals for the same executor, with no significant difference between the two (59 against 62 episodes won), and the remaining failures sit where the executor itself is weak.
InsertPeg, for example, is solved in only 9 of 50 episodes even when the subgoals come from the simulator, and StopCube in 19 (\cref{tab:main} lists our planner per task).

\begin{figure}[tb]
  \centering
  \includegraphics[width=\linewidth]{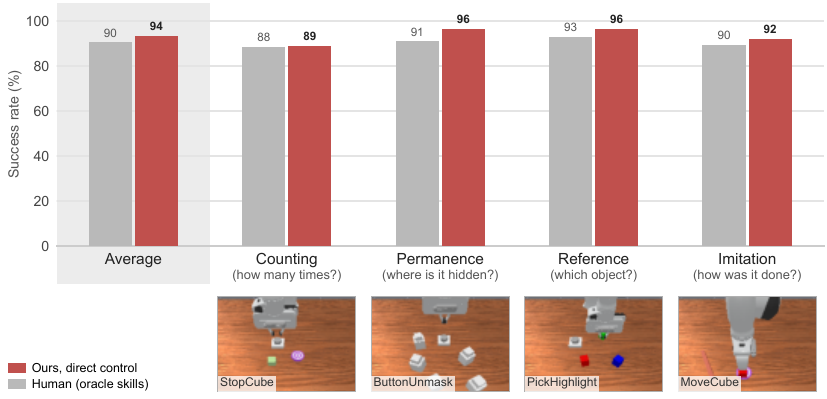}
  \caption{Our planner with direct control on all 800 RoboMME test episodes, averaged and per suite, next to the human performance reported by RoboMME, which was obtained with the oracle skill interface. Shown separately from \cref{fig:results} because the executor differs. Below: frames from our episodes.}
  \label{fig:results_dc}
  \vspace{-0.3cm}
\end{figure}

\paragraph{With direct control.}
The planner solves 93.5\% of all 800 test episodes (\cref{fig:results_dc}).
It is above the oracle-subgoal result of the \pio{} executor in every suite, a gap that comes largely from the more reliable executor, and on the official per-episode results it wins against MemER and FrameSamp on nearly every episode in which they disagree.
Its success varies little with difficulty (\cref{tab:diff}). Of the 52 failed episodes, 35 end in a motion-planning failure, most often in SwingXtimes and MoveCube; StopCube, where a button has to be pressed at the moment a moving cube passes the target, accounts for most of the others (\cref{tab:main}).
In concurrent work, Jedlicki et~al.~\citep{jedlicki2026frontier} let \astra{} command waypoints with the whole episode in context and reach 92.1\%, also counting motion-planning failures as failed episodes.

\begin{table}[tb]
\centering\small
\caption{Task notes with direct control, test episodes 4--19 of every task. Removing the completion criteria keeps the description of each task; removing the notes leaves only the instruction. Wins and losses count the episodes in which only one of the two settings succeeds.}
\label{tab:notes}
\resizebox{\linewidth}{!}{\begin{tabular}{l r r r r}
\toprule
Task notes & Success (\%) & Failures: planning / other & Full notes wins / losses & Calls / ep. \\
\midrule
\rowcolor{oursrow}Ours, full notes & \textbf{237/256 (92.6)} & 15 / 4 & -- & 7.3 \\
Without completion criteria & 236/256 (92.2) & 9 / 11 & 14 / 13 & 8.5 \\
Without task notes & 228/256 (89.1) & 13 / 15 & 19 / 10 & 8.7 \\
\bottomrule
\end{tabular}
}
\vspace{-0.3cm}
\end{table}

\paragraph{Task notes.}
Our notes tell the planner when the benchmark counts a step as done.
We rerun our planner on test episodes 4--19 once with the completion criteria removed from the notes and once without any notes (\cref{tab:notes}).
Without the criteria, success hardly changes, because motion-planning failures again decide most of the episodes in which the two settings disagree; the other failures, however, rise from 4 to 11, ten of them in SwingXtimes and StopCube. In SwingXtimes the planner carries the cube over the targets at 14--16\,cm, whereas a visit counts only below about 12\,cm, a threshold that the full note states.
Without any notes the planner still solves 89.1\% of these episodes, with 15 failures other than motion planning. Without notes our memory also keeps its cost advantage: on the hard episodes it uses about half the tokens of the note-only memory without notes (32k against 59k).
The notes thus help on tasks whose success condition cannot be seen in the images, while most of the planner's success does not depend on them.

\section{Limitations}
All results are from one simulation benchmark and one planner model, and each configuration is run once per episode; repeated runs of the same episode do not always agree, which is why we rely on paired comparisons over many episodes.
Direct control uses depth and camera calibration that the RoboMME learned baselines do not use; we therefore also report the learned-executor setting, with the same executor as those baselines. In a small check with an earlier version of the planner on training episodes, replacing depth by the intersection of each pixel's camera ray with a height estimated by the model gave similar success (27 against 28 of 30 episodes).
The task notes for direct control include completion thresholds taken from the benchmark's success-checking code; without them, the planner fails more episodes for reasons other than motion planning (\cref{sec:full}). Motion-planning failures of the benchmark's planner end 2--12\% of the episodes of every configuration and limit how precisely success rates can separate two memories. The learned-executor variant uses task notes that include conventions of the executor's phrases.
We have not evaluated on a physical robot.
\astra{} is a proprietary model accessed through an API.

\section{Conclusion}
This paper presented execution-grounded memory, a memory for frontier-model robot planners that lets the system keep the record of what was executed and measured while the model decides which frames and which parts of a demonstration to look at.
It addresses two requirements that a note rewritten at every step meets poorly: keeping track of what was done and learned, and choosing what to look at.
On RoboMME, measured with the benchmark's motion planner, the memory reaches the success of a note-only memory at well under half the image cost and with far fewer failures caused by memory and observation, and with direct control the planner solves 93.5\% of the test set; with a learned executor, the executor's own failures decide which episodes are lost.
Future work will evaluate the same memory on a second benchmark, with other planner models and on a physical robot.

{\small
\bibliography{refs}
}

\clearpage
\appendix
\section{Per-task results}

\begin{table}[h]
\centering\small
\caption{Per-task successes on the subsets used in the memory swap (\cref{tab:ablation}); bold marks our planner, not the best entry. Left: learned \pio{} executor, test episodes 4--11 (MemER and FrameSamp: official runs; ``Per-chunk'': the per-chunk planner). Right: memory swap with direct control, test episodes 4--19.}
\label{tab:subsets}
\resizebox{\linewidth}{!}{\begin{tabular}{l r r r r r | r r}
\toprule
 & \multicolumn{5}{c|}{$\pi_{0.5}$ executor, 8 episodes per task} & \multicolumn{2}{c}{Motion planner, 16 ep.} \\
Task & MemER & FrameSamp & Oracle sg. & Per-chunk & Ours & Exec.-gr. & Note-only \\
\midrule
BinFill & 4 & 4 & 6 & 6 & \textbf{7} & \textbf{16} & 15 \\
PickXtimes & 5 & 6 & 8 & 8 & \textbf{8} & \textbf{15} & 14 \\
SwingXtimes & 4 & 8 & 8 & 8 & \textbf{8} & \textbf{14} & 16 \\
StopCube & 0 & 2 & 4 & 1 & \textbf{3} & \textbf{13} & 15 \\
VideoUnmask & 7 & 3 & 7 & 8 & \textbf{8} & \textbf{16} & 16 \\
ButtonUnmask & 4 & 3 & 7 & 8 & \textbf{7} & \textbf{14} & 13 \\
VideoUnmaskSwap & 2 & 1 & 8 & 7 & \textbf{8} & \textbf{14} & 16 \\
ButtonUnmaskSwap & 1 & 2 & 7 & 3 & \textbf{4} & \textbf{16} & 12 \\
PickHighlight & 6 & 1 & 4 & 7 & \textbf{7} & \textbf{15} & 15 \\
VideoRepick & 1 & 2 & 8 & 5 & \textbf{8} & \textbf{15} & 14 \\
VideoPlaceButton & 2 & 4 & 8 & 8 & \textbf{8} & \textbf{16} & 16 \\
VideoPlaceOrder & 2 & 5 & 7 & 8 & \textbf{8} & \textbf{15} & 16 \\
MoveCube & 8 & 8 & 7 & 6 & \textbf{5} & \textbf{13} & 14 \\
InsertPeg & 0 & 1 & 1 & 0 & \textbf{0} & \textbf{15} & 12 \\
PatternLock & 0 & 5 & 8 & 8 & \textbf{8} & \textbf{16} & 16 \\
RouteStick & 1 & 5 & 2 & 3 & \textbf{7} & \textbf{14} & 14 \\
\midrule
Total & 47 & 60 & 100 & 94 & \textbf{104} & \textbf{237} & 234 \\
\bottomrule
\end{tabular}
}
\end{table}

\begin{table}[h]
\centering\small
\caption{Success by difficulty level. Top: all 800 test episodes; MemER and FrameSamp-Modul from the official per-episode runs released with RoboMME; oracle subgoals from our run of RoboMME's evaluation code. Bottom: the memory swap with the motion planner.}
\label{tab:diff}
\begin{tabular}{@{}l r r r r@{}}
\toprule
Difficulty & MemER & FrameSamp-Modul & Oracle subgoals & \textbf{Ours, direct control} \\
\midrule
Easy & 201/416 (48.3) & 241/416 (57.9) & 365/416 (87.7) & \textbf{384/416 (92.3)} \\
Medium & 90/192 (46.9) & 83/192 (43.2) & 153/192 (79.7) & \textbf{187/192 (97.4)} \\
Hard & 50/192 (26.0) & 48/192 (25.0) & 140/192 (72.9) & \textbf{177/192 (92.2)} \\
\midrule
\multicolumn{5}{@{}l}{\textit{Memory swap, motion planner, test episodes 4--19}} \\
Difficulty & \multicolumn{2}{r}{Note-only memory} & \multicolumn{2}{r}{\textbf{Execution-grounded}} \\
Easy & \multicolumn{2}{r}{119/128} & \multicolumn{2}{r}{\textbf{118/128}} \\
Medium & \multicolumn{2}{r}{62/64} & \multicolumn{2}{r}{\textbf{63/64}} \\
Hard & \multicolumn{2}{r}{53/64} & \multicolumn{2}{r}{\textbf{56/64}} \\
\bottomrule
\end{tabular}

\end{table}

\begin{table}[h]
\centering\footnotesize
\caption{Every episode of the memory swap in which exactly one memory succeeds, with the cause of the other memory's failure, read from the logs and checked against the executed-step feedback. A motion-planning failure ends the episode as a failure, as in RoboMME. ``Replay'': the logged commands re-executed without model calls.}
\label{tab:attrib}
\resizebox{\linewidth}{!}{\begin{tabular}{@{}l l l@{}}
\toprule
Episode (task, test ep.) & Succeeds with & Cause of the failure \\
\midrule
\multicolumn{3}{@{}l}{\textit{Motion planner, test episodes 4--19}} \\
BinFill 7 & execution-grounded & count drift: a deposited cube is not counted \\
ButtonUnmaskSwap 19 & execution-grounded & fact dropped: swap tracking lost in a rewrite \\
ButtonUnmaskSwap 5 & execution-grounded & motion planning failure of the other memory (no path found; counted as a failure, as in RoboMME) \\
ButtonUnmaskSwap 6 & execution-grounded & motion planning failure of the other memory (no path found; counted as a failure, as in RoboMME) \\
ButtonUnmaskSwap 9 & execution-grounded & motion planning failure of the other memory (no path found; counted as a failure, as in RoboMME) \\
ButtonUnmask 15 & execution-grounded & fact dropped: colour-to-container mapping lost in a rewrite \\
ButtonUnmask 18 & execution-grounded & motion planning failure of the other memory (no path found; counted as a failure, as in RoboMME) \\
InsertPeg 11 & execution-grounded & execution: grasp orientation the arm cannot rotate out of \\
InsertPeg 13 & execution-grounded & demonstration misread: wrong peg \\
InsertPeg 19 & execution-grounded & demonstration misread: wrong end of the peg \\
InsertPeg 8 & execution-grounded & demonstration misread: wrong peg \\
MoveCube 7 & execution-grounded & demonstration misread: push read as carry \\
PickHighlight 7 & execution-grounded & fact dropped: failed grasp noted as held, target later dropped from the note \\
PickXtimes 11 & execution-grounded & count drift: a completed placement is announced again \\
PickXtimes 15 & execution-grounded & count drift \\
RouteStick 16 & execution-grounded & motion planning failure of the other memory (no path found; counted as a failure, as in RoboMME) \\
RouteStick 4 & execution-grounded & demonstration misread: route reversed \\
VideoRepick 19 & execution-grounded & count drift \\
VideoRepick 8 & execution-grounded & count drift \\
ButtonUnmask 4 & note-only & motion planning failure of the other memory (no path found; counted as a failure, as in RoboMME) \\
InsertPeg 14 & note-only & motion planning failure of the other memory (no path found; counted as a failure, as in RoboMME) \\
MoveCube 13 & note-only & demonstration misread by our planner (carry), despite a closer look \\
MoveCube 4 & note-only & motion planning failure of the other memory (no path found; counted as a failure, as in RoboMME) \\
PickHighlight 11 & note-only & motion planning failure of the other memory (no path found; counted as a failure, as in RoboMME) \\
PickXtimes 5 & note-only & motion planning failure of the other memory (no path found; counted as a failure, as in RoboMME) \\
RouteStick 19 & note-only & motion planning failure of the other memory (no path found; counted as a failure, as in RoboMME) \\
RouteStick 7 & note-only & motion planning failure of the other memory (no path found; counted as a failure, as in RoboMME) \\
StopCube 11 & note-only & motion planning failure of the other memory (no path found; counted as a failure, as in RoboMME) \\
StopCube 5 & note-only & timing: StopCube press too late \\
SwingXtimes 17 & note-only & motion planning failure of the other memory (no path found; counted as a failure, as in RoboMME) \\
SwingXtimes 4 & note-only & motion planning failure of the other memory (no path found; counted as a failure, as in RoboMME) \\
VideoPlaceOrder 15 & note-only & motion planning failure of the other memory (no path found; counted as a failure, as in RoboMME) \\
VideoRepick 7 & note-only & motion planning failure of the other memory (no path found; counted as a failure, as in RoboMME) \\
VideoUnmaskSwap 4 & note-only & execution: grasp slipped \\
VideoUnmaskSwap 7 & note-only & motion planning failure of the other memory (no path found; counted as a failure, as in RoboMME) \\
\midrule
\multicolumn{3}{@{}l}{\textit{\pio{} executor, test episodes 4--11}} \\
BinFill 10 & execution-grounded & executor: stall until the step limit \\
BinFill 6 & execution-grounded & executor: empty grasp \\
MoveCube 7 & execution-grounded & demonstration misread by note-only memory: push read as carry \\
PickHighlight 7 & execution-grounded & executor: empty grasp \\
RouteStick 5 & execution-grounded & executor: identical command replay succeeds \\
RouteStick 7 & execution-grounded & executor: identical command replay succeeds \\
StopCube 9 & execution-grounded & executor: press latency \\
BinFill 11 & note-only & executor: stall until the step limit \\
ButtonUnmaskSwap 10 & note-only & executor: button press not registered \\
ButtonUnmask 11 & note-only & executor: empty grasp \\
MoveCube 11 & note-only & executor: stall \\
MoveCube 8 & note-only & executor: stall \\
PickHighlight 11 & note-only & undetected grasp of the wrong cube \\
StopCube 6 & note-only & executor: press latency \\
\bottomrule
\end{tabular}
}
\end{table}

\clearpage
\section{Tasks, an example episode and failure cases}

\begin{figure}[h]
\centering
\includegraphics[width=\linewidth]{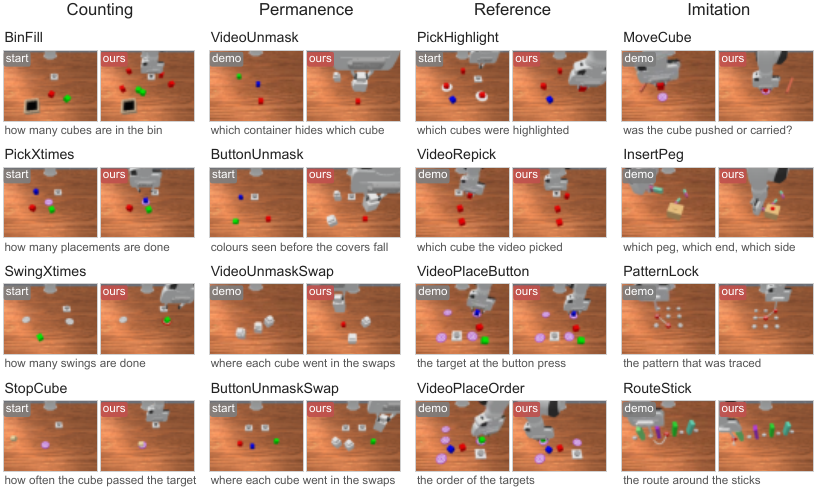}
\caption{The 16 RoboMME tasks. For each task, the left frame shows what has to be remembered (a frame of the episode's demonstration video, or the scene at the first decision for tasks without a video), and the right frame is taken from a successful episode of our planner with direct control. The line below names what the planner has to keep track of.}
\label{fig:gallery}
\end{figure}

\begin{figure}[h]
\centering
\includegraphics[width=\linewidth]{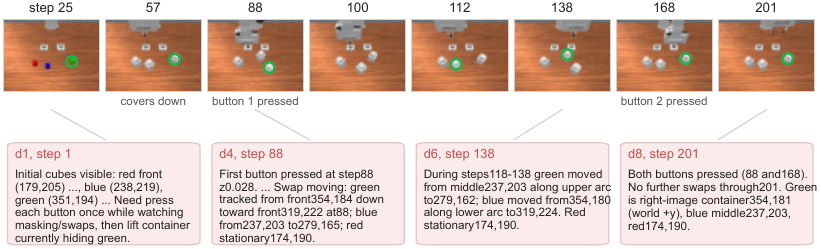}
\caption{Selected frames and notes from one episode of our planner with direct control (ButtonUnmaskSwap, test episode 8, success). Top: front frames at the given environment steps, from an exact replay of the logged commands; the green circle marks the container that the model's note says hides the green cube. Bottom: four of the nine notes, quoted verbatim (positions in the 512-pixel image). The model watched the containers during each move; after the six watched actions the system returned 14, 7, 6, 7, 8 and 9 of the 79 recorded frames.}
\label{fig:timeline}
\end{figure}

\begin{figure}[h]
\centering
\includegraphics[width=\linewidth]{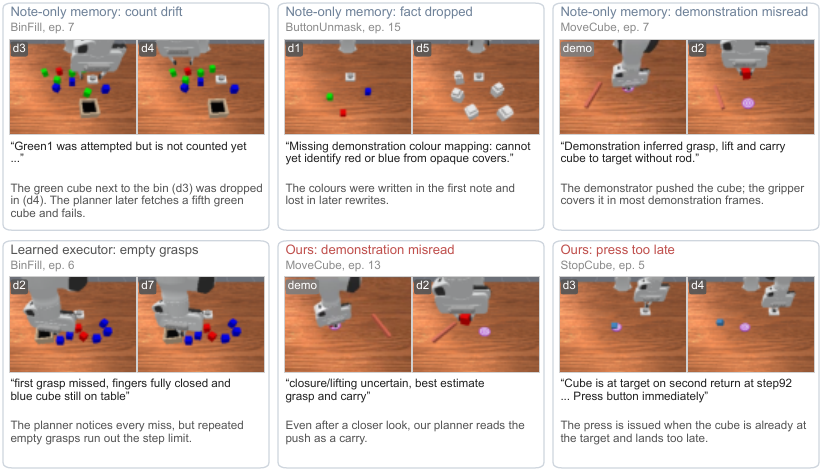}
\caption{Examples of the failure causes in \cref{tab:attrib}. Each panel shows two frames of the failed episode (d$N$: the scene at the $N$-th decision; demo: a frame of the demonstration) and an excerpt of the planner's note, quoted verbatim. Top: the note-only memory with direct control, in episodes that our memory solves. Bottom: a failure of the learned executor, and two failures of our planner with direct control.}
\label{fig:failures}
\end{figure}

\clearpage
\section{Implementation details}\label{app:impl}
\Cref{tab:impl} lists the settings of our planner and of the note-only memory. All of them are the same for every task.

\begin{table}[h]
\centering\small
\caption{Implementation details. All values are identical for every task; the note-only memory differs from ours only in the last block.}
\label{tab:impl}
\resizebox{\linewidth}{!}{\begin{tabular}{@{}l l@{}}
\toprule
Parameter & Value \\
\midrule
\multicolumn{2}{@{}l}{\textit{Model and inputs}} \\
Planner model & \astra{} through its API, reasoning effort ``high'' \\
Maximum output per call & 6000 tokens \\
Images at each decision & front and wrist camera, $512\times512$ \\
Demonstration frames & $256\times256$ \\
\midrule
\multicolumn{2}{@{}l}{\textit{Budgets per episode}} \\
Environment steps & 1300 (benchmark limit) \\
Model decisions & at most 80 \\
Tokens & 1.5\,M \\
\midrule
\multicolumn{2}{@{}l}{\textit{Actions}} \\
Sequence & at most 6 moves or waits, optional check on gripper opening or height \\
\pio{} delegation & 16-step action chunks, $k \in \{1,\dots,8\}$ chunks per call \\
Executors & RoboMME motion planner (waypoints); RoboMME \pio{} grounded-subgoal checkpoint \\
\midrule
\multicolumn{2}{@{}l}{\textit{Execution-grounded memory}} \\
Execution log & all executed actions; command, environment step, finger opening, end-effector height \\
Note history & all earlier notes, verbatim \\
Frames recorded during an action & every 4 environment steps, at most 16 per action \\
Watch points & at most 8 per action, or ``all'' \\
Change test & radius 28\,px (at 512\,px); $>$5\% of pixels in the circles change by $>$30 levels \\
Returned frames & each changed frame and its predecessor \\
Demonstration skim & every 8th frame, segments of 24 shown frames \\
Closer look & every 2nd frame of the requested ranges, at most 24 frames per segment \\
\midrule
\multicolumn{2}{@{}l}{\textit{Note-only memory (baseline)}} \\
Record of actions & last 8 executed actions \\
Frames during an action & all recorded frames \\
Demonstration & every 2nd frame (every 4th above 200 frames), segments of 24 \\
\bottomrule
\end{tabular}
}
\end{table}

\end{document}